%% file: main.tex
\documentclass[sigconf]{aamas}

\usepackage{balance} % for balancing columns on the final page
\usepackage{xcolor}
\usepackage{colortbl}
\usepackage{subcaption} % for subfigures
\usepackage{algorithm} % for algorithms
\usepackage{algpseudocode}
\usepackage{amsmath} % for mathematical notation
\definecolor{mytheoremfr}{RGB}{122, 106, 226} % Frame color
\definecolor{mytheorembg}{RGB}{224, 214, 250} % Background color

\usepackage[most,skins,theorems]{tcolorbox}
\tcbset{
  aibox/.style={
    width=\linewidth,
    top=8pt,
    bottom=4pt,
    colback=blue!6!white,
    colframe=black,
    colbacktitle=black,
    enhanced,
    center,
    attach boxed title to top left={yshift=-0.1in,xshift=0.15in},
    boxed title style={boxrule=0pt,colframe=white,},
  }
}

\newtcolorbox{AIbox}[2][]{aibox,title=#2,#1}

\newtcbtheorem{Prompt}{\textbf{Prompt}}{colback=mytheorembg!25,colframe=mytheoremfr, boxrule=0.5pt
}{th}

\newcommand{\legomem}{LEGOMem}
\newcommand{\dynalegomem}{LEGOMem-Dynamic}
\newcommand{\queryrewritelegomem}{LEGOMem-QueryRewrite}

\setcopyright{ifaamas}
\acmConference[AAMAS '26]{Proc.\@ of the 25th International Conference
on Autonomous Agents and Multiagent Systems (AAMAS 2026)}{May 25 -- 29, 2026}
{Paphos, Cyprus}{C.~Amato, L.~Dennis, V.~Mascardi, J.~Thangarajah (eds.)}
\copyrightyear{2026}
\acmYear{2026}
\acmDOI{}
\acmPrice{}
\acmISBN{}

\acmSubmissionID{14}

\title[\legomem{}: Modular Procedural Memory]{\legomem{}: Modular Procedural Memory for Multi-agent LLM Systems for Workflow Automation}

\author{Dongge Han, Camille Couturier, Daniel Madrigal Diaz, Xuchao Zhang, \\Victor Rühle, Saravan Rajmohan}\thanks{correspond to Dongge Han at donggehan@microsoft.com}
\affiliation{
  \institution{Microsoft}
  \country{}
}
\email{}

\begin{abstract}
We introduce \legomem{}, a modular procedural memory framework for multi-agent large language model (LLM) systems in workflow automation. \legomem{} decomposes past task trajectories into reusable memory units and flexibly allocates them across orchestrators and task agents to support planning and execution. To explore the design space of memory in multi-agent systems, we use \legomem{} as a lens and conduct a systematic study of procedural memory in multi-agent systems, examining where memory should be placed, how it should be retrieved, and which agents benefit most. Experiments on the OfficeBench benchmark show that orchestrator memory is critical for effective task decomposition and delegation, while fine-grained agent memory improves execution accuracy. We find that even teams composed of smaller language models can benefit substantially from procedural memory, narrowing the performance gap with stronger agents by leveraging prior execution traces for more accurate planning and tool use.
These results position \legomem{} as both a practical framework for memory-augmented agent systems and a research tool for studying memory design in multi-agent workflow automation.
\end{abstract}
\ccsdesc[500]{Computing methodologies~Multi-agent systems}
\ccsdesc[300]{Computing methodologies~Reasoning about beliefs and knowledge}

\keywords{Multi-agent systems, Procedural memory, LLM Agents, Workflow}

\newcommand{\BibTeX}{\rm B\kern-.05em{\sc i\kern-.025em b}\kern-.08em\TeX}

\begin{document}

%%% The following commands remove the headers in your paper. For final 
%%% papers, these will be inserted during the pagination process.

\pagestyle{fancy}
\fancyhead{}

%%% The next command prints the information defined in the preamble.

\maketitle 

%%%%%%%%%%%%%%%%%%%%%%%%%%%%%%%%%%%%%%%%%%%%%%%%%%%%%%%%%%%%%%%%%%%%%%%%

\section{Introduction}
% Outline
% 1. LLM models for workflow automation, and the importance of multi-agent LLM systems in such settings,
% 2. then, mention the lack of memory, especially for complex workflow scenarios, lack of procedural memory harms agents and make them highly transactional, (motivates procedural memory, and lack of such memory systems for multi-agent systems), here we can start to cite a few baselines like awm, synapse, which are for single-agent systems, and then a few other memory papers that are not procedural. 
% 3. next paragraph, we talk about how our work aims to address this, i.e., the motivation of our legomem, what system we expected to build and how we did build this to address the existing challenges. 
% 4. then a bit of more details in how our legomem's logic is, and why is it good. 
% 5. then evaluations, the benchmark, and the main insights we get from there.

%% LEGOMem OVERVIEW FIGURE
\begin{figure*}
    \centering
    \begin{subfigure}{0.61\textwidth}
        \centering
        \includegraphics[width=\linewidth]{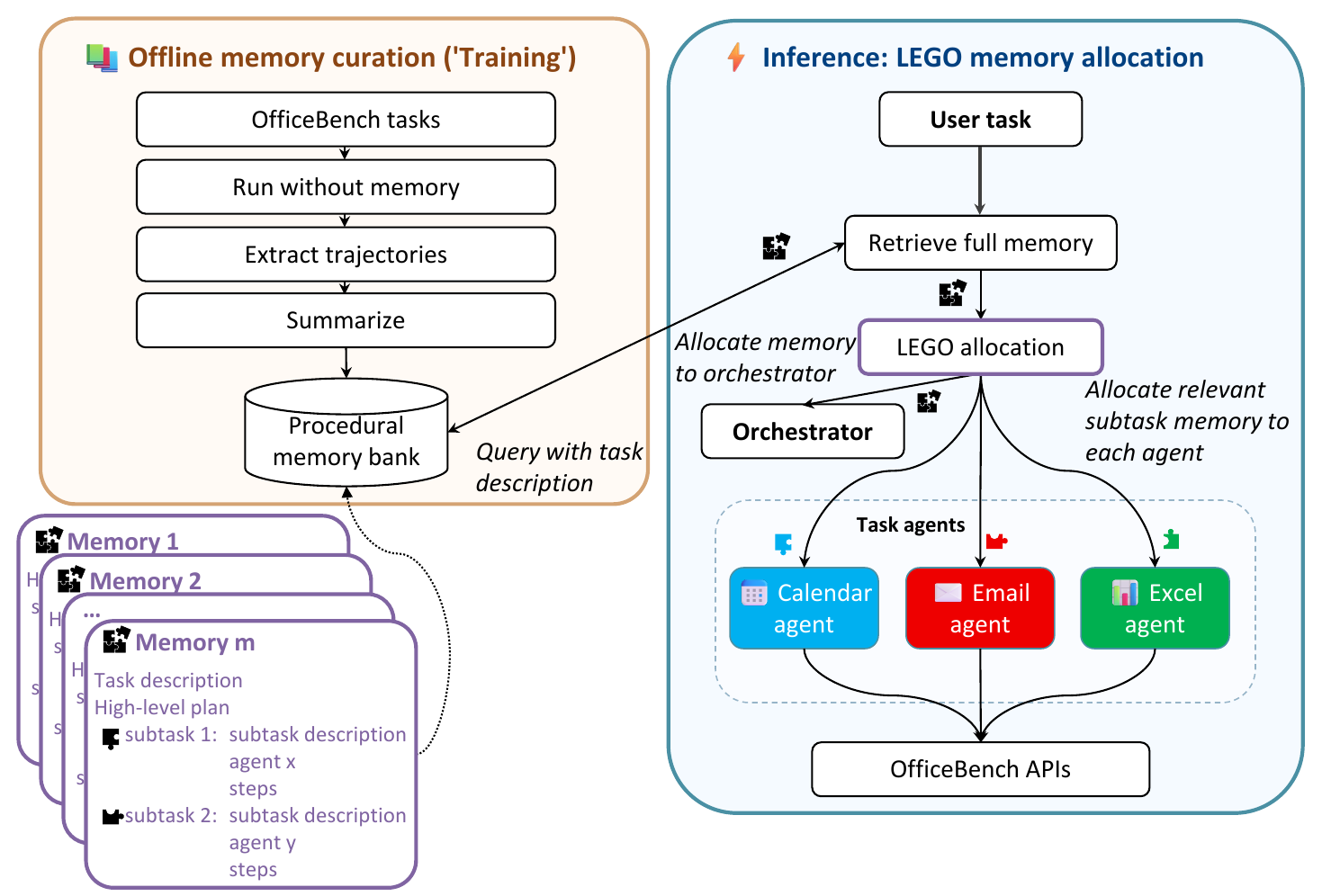}
        \caption{Overview of the LEGOMem{} framework}
        \label{fig:framework-overview-sub}
    \end{subfigure}
    \hfill
    \begin{subfigure}{0.38\textwidth}
        \centering
        \includegraphics[width=\linewidth]{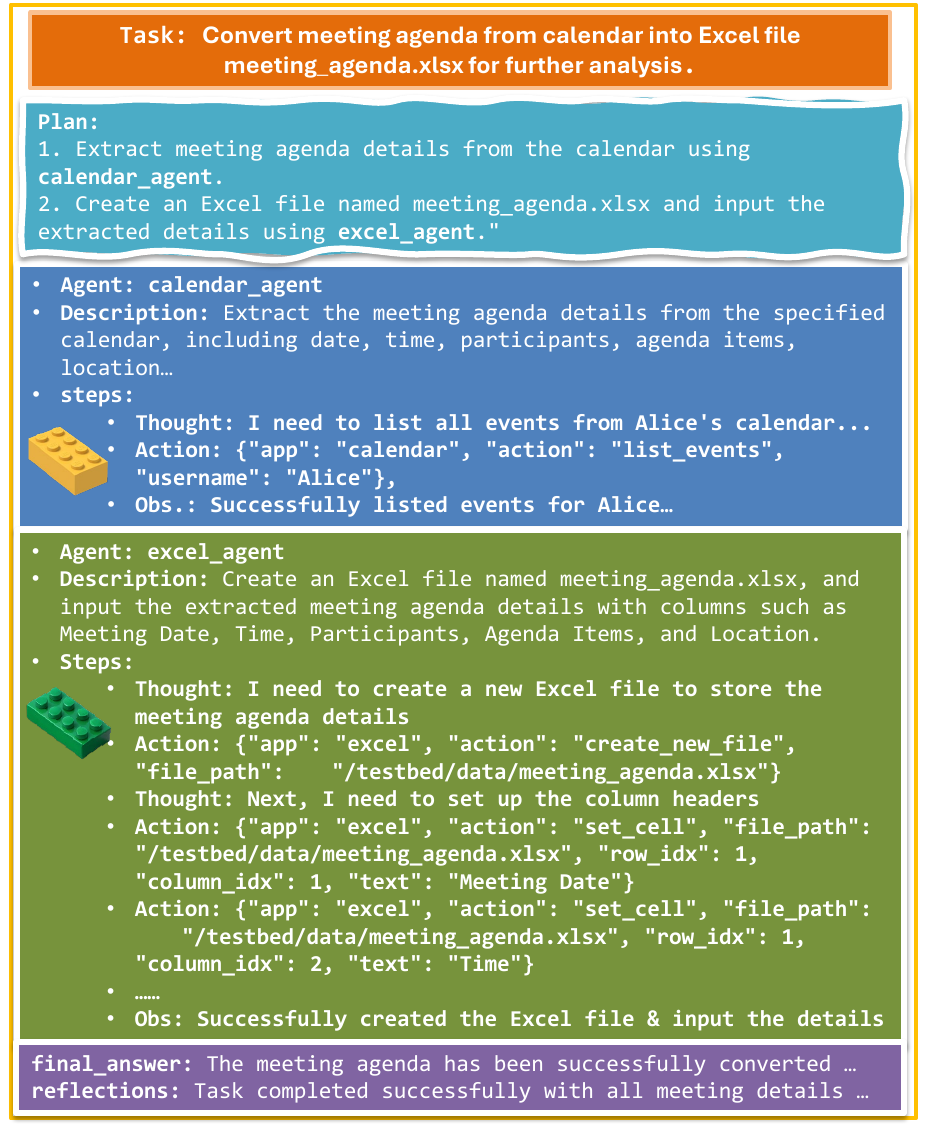}
        \caption{Example \legomem{} memory structure}
        \label{fig:memory-example}
    \end{subfigure}
    \caption{\legomem{} framework overview and example memory. The multi-agent system consists of an orchestrator and task agents. The orchestrator performs planning, next agent selection, and subtask allocation, while task agents execute subtasks by interacting with the environment via API tool calls. \textmd{(Note: For clarity, additional task agents such as Word agent are omitted.)}
    % LEGOMem ...
    }
    \label{fig:legomem-framework}
\end{figure*}

Large Language Models (LLMs) are increasingly deployed as agents to automate complex multi-step workflows~\cite{yao2023react, song2023llm, wang2023plan, kim2024llm, wang2024officebench, xie2024osworld, wang2025odysseybench, zhou2023webarena, mialon2023gaia, cao2024spider2, rana2023sayplan, ahn2022can, wang2024survey, cheng2024exploring}. These agents are especially valuable in productivity environments such as document editing, email handling, and calendar scheduling. To manage the diversity and compositionality of such tasks, recent systems often adopt multi-agent~\cite{stone2000multiagent, 10.5555/1695886} designs, where multiple LLM-based agents collaborate, specialize, or delegate responsibilities across roles and tools~\cite{fourney2024magentic, wu2024autogen, dang2025multi, chen2023agentverse, zhang2024ufo}. This trend reflects a broader shift in AI system design: the real world is inherently multi-agent, involving heterogeneous roles and coordinated decision-making. Multi-agent LLM systems offer a scalable and modular approach to reasoning, tool-use, and workflow execution, positioning them as a natural fit for these increasingly complex productivity environments.

Despite these advances, current multi-agent systems remain largely stateless and transactional: each task is solved from scratch, without reusing prior experience. This lack of memory—particularly procedural memory—limits their ability to learn from past experiences and build up execution skills over time for complex workflows. While recent works have proposed memory modules for single-agent LLMs, such as Synapse~\cite{zheng2023synapse}, Agent Workflow Memory (AWM)~\cite{wangAgentWorkflowMemory2024}, these approaches do not address the unique coordination and specialization challenges of multi-agent systems. 

To address this gap, we introduce \legomem{}, a modular procedural memory framework designed for multi-agent LLM systems. In this work, we focus on a common and practical subclass of multi-agent architectures, where a central orchestrator performs planning and delegates subtasks to specialized tool-using task agents, as exemplified by the Magentic-One framework~\cite{fourney2024magentic, wu2024autogen}. Our goal is to equip both orchestrators and task agents with memory grounded in prior task trajectories, enabling them to perform better planning, coordination, and task executions. 
To this end, we design \legomem{} to distill successful executions into structured memory units: full-task memories (task-level plans and reasoning traces) and subtask memories (agent behavior and tool interactions). These modular memories are stored in a memory bank, indexed by semantic embeddings, and reused at inference time to augment planning and execution.

\legomem{} is instantiated as a retrieval augmentation (RAG)~\cite{lewis2020retrieval, gao2023retrieval, douze2024faiss} layer over existing multi-agent systems. During a new task, the orchestrator receives relevant full-task memories to support task decomposition and agent selection, while each task agent is assigned subtask memories aligned with its delegated subtasks. We explore three memory retrieval strategies—\textit{vanilla}, \textit{dynamic retrieval}, and \textit{query rewriting}—to study how retrieval and memory specificity affect multi-agent performance.
This framework allows us to systematically investigate key questions in multi-agent memory design, including where memory should be placed, how it should be retrieved, and which agents benefit most from it.

We evaluate \legomem{} in the context of productivity workflow automation using the OfficeBench~\cite{wang2024officebench} benchmark, with agent teams composed of LLM-only, hybrid, and small language model configurations. Across these settings, all \legomem{} variants significantly improve task success rates over memory-less and baseline methods. Our ablation studies reveal that orchestrator memory is critical for high-level planning and delegation, while fine-grained subtask retrieval provides meaningful gains for smaller agents that rely more on localized execution support. These findings highlight how memory placement and retrieval strategy shape the effectiveness of multi-agent collaboration in workflow settings.
Overall, \legomem{} provides a practical and extensible framework for memory-augmented multi-agent workflow automation, enabling agents to plan, coordinate, and execute more effectively by reusing structured procedural knowledge. We hope this work facilitates further research on memory design, continual learning, and efficient agent collaboration in complex productivity settings.

\section{Related work}
\paragraph{Multi-agent LLM systems for workflow automation} 
The recent advent of LLMs has enabled the development of multi-agent systems able to plan, decompose and solve complex workflows.
Generalist multi-agent frameworks~\cite{fourney2024magentic, wu2024autogen, dang2025multi, chen2023agentverse, zhang2024ufo} such as Magentic-One~\cite{fourney2024magentic} use a common design pattern where a lead orchestrator agent decomposes high-level goals into a step-by-step plan and directs a team of specialized agents to execute specific subtasks. This modular, multi-agent architecture simplifies development and facilitates the reuse of encapsulated skills, a significant advantage over monolithic, single-agent approaches.
However, a key limitation remains that they are often stateless, solving each task from scratch and discarding valuable insights gained during execution. Without memory, agents may repeatedly make the same errors and cannot improve over time.

\paragraph{Memory for LLM agents}
Memory offers a natural solution to the limitations of stateless agents. However, a primary challenge is that memory in LLM agents is often designed for single-agent systems and are often episodic/semantic, replaying information from dialogue histories~\cite{zhong2024memorybank, sun2025hierarchical, rasmussen2025zep, wu2024longmemeval, maharana2024evaluating}, such as A-MEM~\cite{xuAMEMAgenticMemory2025} which captures interactions as a network of interconnected notes that form an evolving memory structure, and Mem0~\cite{chhikaraMem0BuildingProductionReady2025}, which focus on managing memory from ongoing conversations.
While these systems advance memory capabilities, they are not designed for agentic learning and workflow automation.
Another line of works target memory optimization for agentic workflows including~\cite{lee2025learning, kang2025acon, zhou2025mem1}, which focus on short-term context optimization for workflow agents. 
Most closely related works on long-term, procedural memory for agents include Synapse~\cite{zheng2023synapse}, which uses successful past full trajectories as exemplars, and Agent Workflow Memory (AWM)~\cite{wangAgentWorkflowMemory2024} which induces frequently used subtask sequences as reusable skills. However, both works target procedural memory for single-agent scenarios. In contrast, \legomem{} introduces modular, role-aware procedural memory for multi-agent systems. By flexibly allocating memory across orchestrators and task agents, it addresses unique challenges in memory placement and allocation, improving workflow automation through better planning, execution, and coordination.

\section{\legomem{}: Modular Procedural Memory for Multi-agent LLM Systems}
\label{sec:method}
In this section, we introduce \legomem{}, a modular procedural memory framework designed for multi-agent LLM systems. We begin by formalizing the problem setting of multi-agent workflow execution with procedural memory, then present the detailed \legomem{} framework, its variants and the design choices studied in our experiments.

\subsection{Problem formulation}
% the problem formulation will cover the multi-agent setup
% the goal we want to achieve
% the abstract framework we want to implement/achieve with LEGOMem
% Highlight the need for multi-agent systems, which motivates the design considerations for the LegoMem

\subsubsection{Multi-agent system for workflow automation.}
We consider a common multi-agent workflow automation framework (based on Magentic-One system~\cite{fourney2024magentic}) with an orchestrator $A_{\text{orch}}$, a set of task agents $A = \{A_1, \ldots, A_k\}$, and an external environment $\mathcal{E}$. A task $T$ is specified by a natural language description $d$ and must be executed within $\mathcal{E}$. Specifically, we implemented task agents for Word, Excel, Calendar, Email, System, and OCR-PDF apps. These task agents interact with the simulated apps in a Docker environment via tool APIs, ensuring isolated and reproducible execution.

The orchestrator first generates an initial high-level plan $\pi_0 = \{s_1, \ldots, s_m\}$, outlining a possible sequence of subtasks. 
However, orchestration is not a static plan-following process: after each orchestration step, the orchestrator dynamically generates the next subtask based on the current state $\sigma_t$ and observations returned from the agents, rather than simply selecting from the initial plan. 

Formally, at each orchestration step $t$:
\begin{enumerate}
    \item the orchestrator proposes the next subtask $s_t = \pi_{\text{orch}}(\sigma_t)$;
    \item the subtask is assigned to an appropriate task agent $A_j$;
    \item the task agent executes $s_t$ by issuing tool-use commands to the environment $\mathcal{E}$, returning an observation $o_t$ and an execution summary $r_t$;
    \item the orchestrator updates its state $\sigma_{t+1} = f(\sigma_t, r_t)$ and continues orchestration.
\end{enumerate}

If progress stalls (e.g., repeated states or looping behavior), the orchestrator may perform \emph{re-planning}, generating a revised high-level plan $\pi'$, and resuming orchestration from the updated state. 
The system is considered successful if the final environment state $\sigma_{\text{final}} \in \mathcal{E}$ satisfies the task goal.

\subsubsection{Multi-agent procedural memory.} 
% Here we define the motivation of multi-agent procedural memory, and the specific benefits they can bring.
While the orchestration loop above defines how agents interact with the environment, it remains \emph{stateless}: each new task $T$ is solved from scratch, discarding knowledge from past executions. 
To address this limitation and enable agents to improve through experience, we introduce \emph{multi-agent procedural memory}: modular, role-aware memories distilled from successful trajectories and reused across tasks in a multi-agent system. 
In contrast to episodic or semantic memory, which primarily capture events or textual information, multi-agent procedural memory abstracts workflows into reusable subroutines tailored to both orchestrators and task agents. 
These memories allow orchestrators to plan more effectively and select agents with greater context, while equipping task agents with execution-level guidance for more accurate and efficient tool use.

Formally, we define a memory store $M$ as a collection of modular memory units derived from past executions. 
These include \textbf{full memories} that capture orchestration plans and summarized execution traces, as well as \textbf{subtask memories} that capture agent-specific subtask executions. 
Together, they form a role-aware memory library that can be retrieved and allocated during inference to augment both planning and execution.

In the following section, we present \legomem{}, a concrete framework that implements this formulation through structured memory construction, inference-time allocation, and variant strategies for retrieval and memory reuse for more robust workflow automation.

\subsection{The \legomem{} framework}
\label{sec:legomem-framework}

The \legomem{} framework instantiates the problem formulation by equipping multi-agent systems with modular procedural memory. 
It operates in two phases: (i) an offline \emph{memory construction} phase, where successful task trajectories are distilled into reusable memory units; and (ii) an online \emph{memory-augmented inference} phase, where retrieved memories are allocated to the orchestrator and task agents to guide planning and execution. 
As illustrated in \autoref{fig:framework-overview-sub}, past task trajectories are curated into a procedural memory bank, which is then queried at inference time to provide high-level orchestration guidance and agent-specific execution traces. 
\autoref{fig:memory-example} further shows the structure of these memory units, consisting of a high-level plan, localized agent subtask traces, the final answer, and a brief reflection. 
This modular design enables LEGO-like recombination of past experiences to support efficient and reliable task completion across diverse multi-agent environments.

\subsubsection{Memory construction.}

The first phase of \legomem{} is offline memory construction, where successful task trajectories are distilled into structured and reusable memory units. 
From each trajectory, we extract two complementary types of memory: 
(i) \emph{full-task memories} that capture the task description, the high-level plan executed, and  
(ii) \emph{subtask memories}, that encapsulate the subtask description, the localized agent behavior and tool-use, and observations.  
These modular units are stored in a procedural memory bank $\mathcal{M}$ for future reuse. 
At inference time, the orchestrator receives the full-task memory in its entirety, while task agents are provided with the relevant subtask memories.  

Concretely, the construction process operates on execution logs of successfully completed tasks. Each log records the task description, the orchestrator’s planning and orchestration steps, the subtasks delegated to agents, and the corresponding agent executions (tool-use commands, observations, and outcomes). 
We use an LLM to transform these logs into structured \legomem{} units, as shown in \autoref{fig:memory-example}.  
The resulting memory bank $\mathcal{M}$ is implemented as a vector database, indexed using dense embeddings. 
Let $\phi(\cdot)$ denote the embedding model used for indexing; for full-task memories, we compute $\phi(d)$ based on the task description $d$ to enable semantic similarity retrieval.  
We implement and compare three retrieval and allocation strategies: vanilla \legomem{}, \dynalegomem{}, and \queryrewritelegomem{}.  
For vanilla \legomem{}, the entire memory bank $\mathcal{M}$ is indexed using $\phi(d)$, enabling direct retrieval of relevant full-task memories at inference time and subtask allocation to task agents. For the other two variants, we separate the global memory bank $\mathcal{M}$ of full task memories, and subtask memory banks $\{\mathcal{M}_{A_j} | A_j\in A\}$ per task agent, which contain the subtask memories that are easily extracted from the global memories, and subtask memory banks are indexed by the embeddings of the subtask descriptions $\phi(d_\textnormal{subtask})$. More details regarding the \legomem{} variants will be discussed in Section~\ref{sec:legomem-variants}.

\begin{algorithm}[t]
\caption{Multi-agent Execution with Vanilla \legomem{}}
\label{alg:vanilla-legomem}
\begin{algorithmic}[1]
\State \textbf{Input:} task description $d_\text{new}$, memory bank $\mathcal{M}$, orchestrator $A_\textnormal{orch}$, task agents $A=\{A_1, ... A_k\}$
\State Compute embedding of $\phi({d_\text{new}})$ and retrieve top-$K$ semantically similar full-task memories $m=\{m_1, ... , m_K\}$ from $\mathcal{M}$.
\State Extract subtask memories $\{m^\textnormal{1}_1, ... , m^\textnormal{K}_n\}$ from the full-task memories and assign subtask memories corresponding to each agent.
\State Initialize environment $\mathcal{E}$ and start task $d_\text{new}$.
\State Augment retrieved full-task memories $m$ to the orchestrator, which then generates initial plan $\pi_o$.
\While{task not completed}
  \State Orchestrator $A_\textnormal{orch}$ selects next agent $A_t\in A$, generates the next subtask $s_t$ and assign to $A_t$,
  \State Augment subtask memories to the task agent $A_t$
  \State Task agent $A_t$ generates a list of tool-use actions, which are executed in the environment.
  \State Agent receives observation $o_t$, summarize subtask execution and sends summary message $r_t$ to orchestrator $A_\textnormal{orch}$
  \If{progress stalls}
    \State Orchestrator performs re-planning and update plan $\pi'$
  \EndIf
\EndWhile
\State \textbf{return} orchestrator final response.
\end{algorithmic}
\end{algorithm}

%% LEGOMem DYNAMIC AND QUERY-REWRITE FIGURE
\begin{figure*}[t]
    \centering
    \begin{subfigure}{0.48\textwidth}
        \centering
        \includegraphics[width=\linewidth]{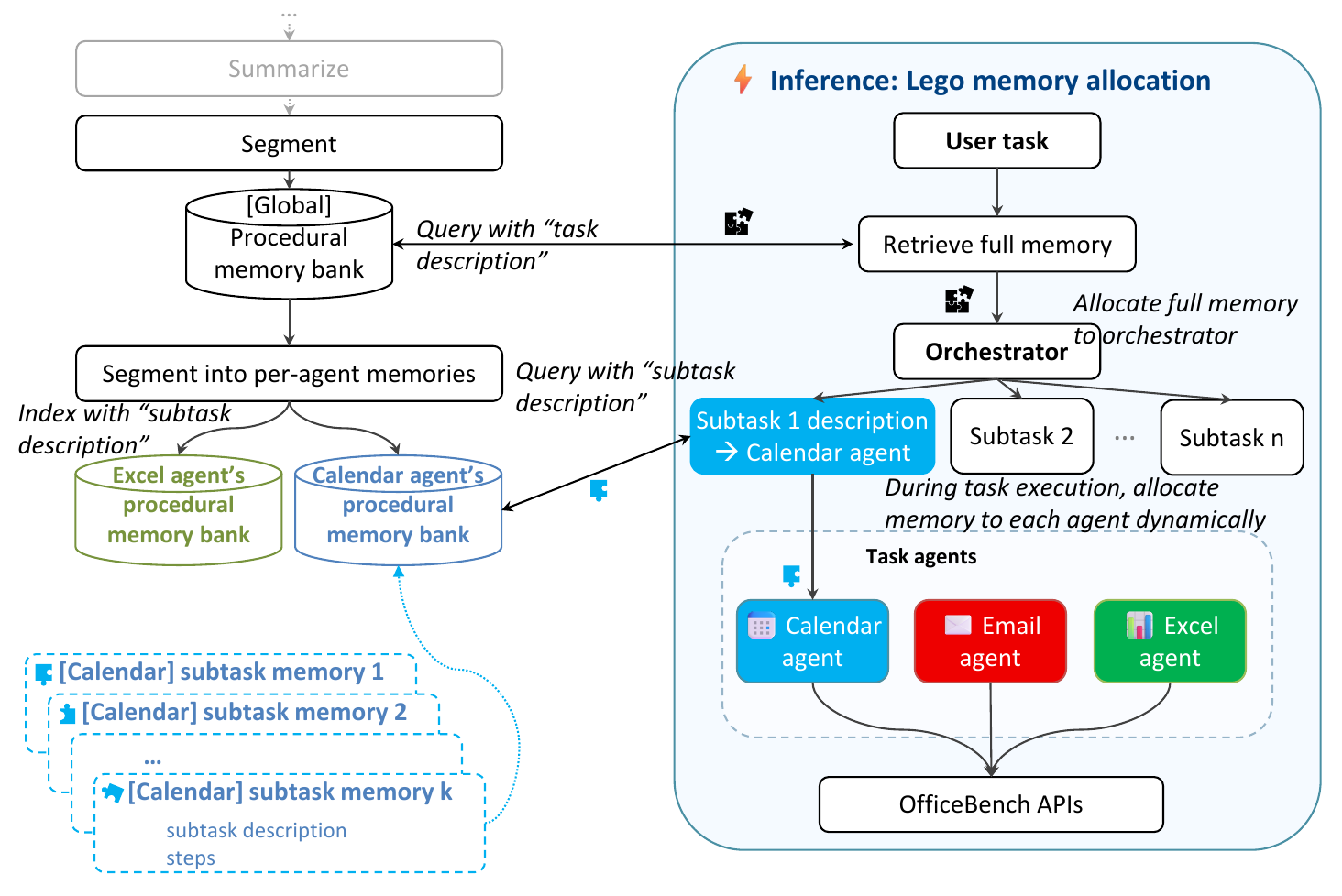}
        \caption{\dynalegomem{} variant}
        \label{fig:dynamic-variant}
    \end{subfigure}
    \hfill
    \begin{subfigure}{0.48\textwidth}
        \centering
        \includegraphics[width=\linewidth]{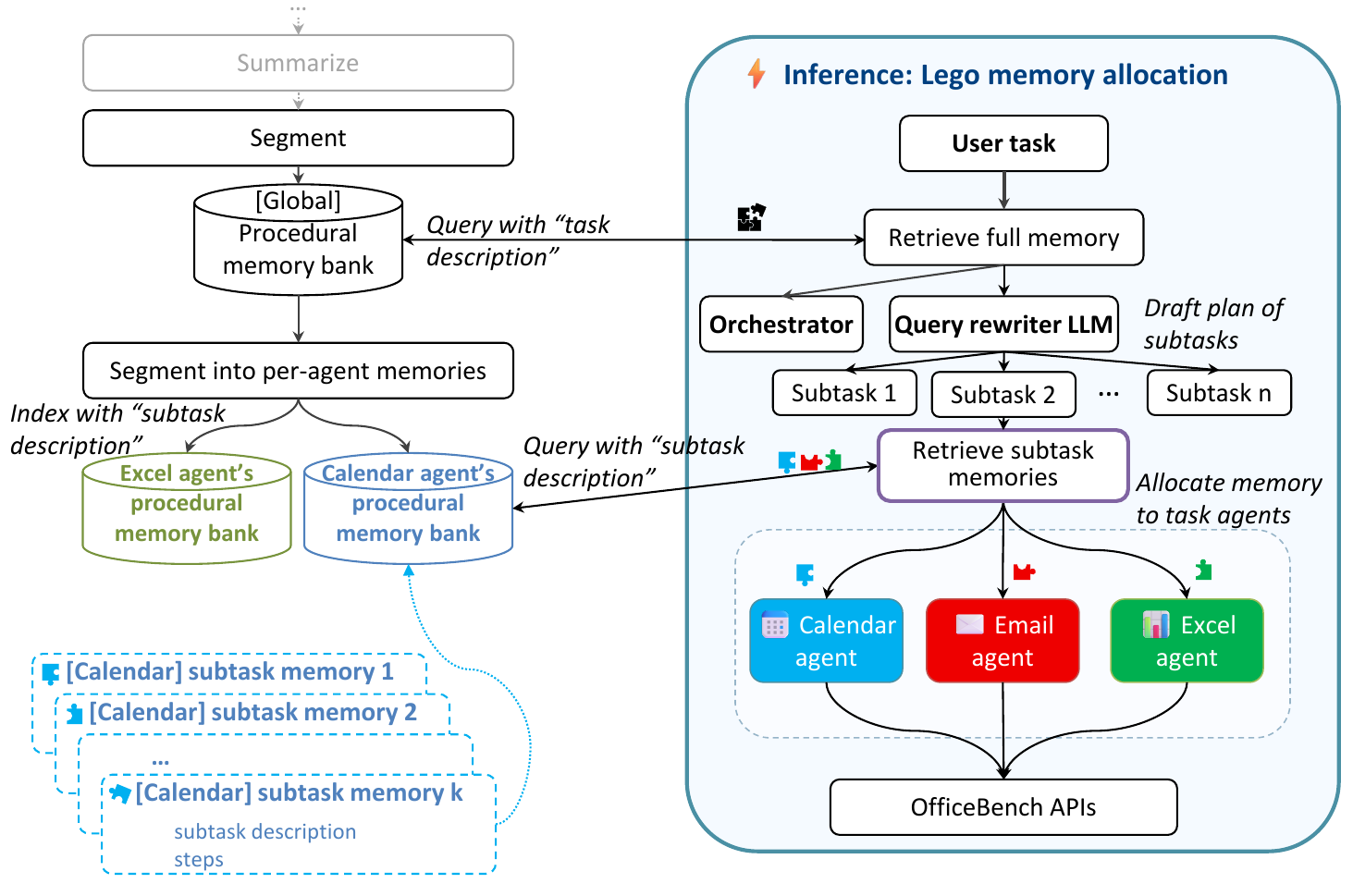}
        \caption{\queryrewritelegomem{} variant}
        \label{fig:query-rewrite-variant}
    \end{subfigure}
    \caption{Comparison of \legomem{} variants: (a) \dynalegomem{} dynamically retrieves subtask memories during execution, and (b) \queryrewritelegomem{} employs query rewriting to retrieve multiple candidate memories for each subtask.}
    \label{fig:legomem-variants}
\end{figure*}

\subsubsection{Memory-augmented inference.}

In the second phase, \legomem{} augments the task execution loop by supplying the \emph{orchestrator} with full-task memories (end-to-end for planning and detailed orchestration) and augment \emph{task agents} with subtask memories (localized execution guidance). Given a new task $d_{\text{new}}$, the system retrieves relevant memories from the memory banks and allocates them accordingly. We designed and tested three different \legomem{} variants which exhibit different memory retrieval strategies, which will be detailed in Section~\ref{sec:legomem-variants}. Here we will describe the vanilla \legomem{} inference approach, as shown in Algorithm~\ref{alg:vanilla-legomem}.

Given a new task with description $d_{\text{new}}$, we obtain the embedding $\phi(d_\text{new})$ and the system first retrieves top-K relevant memories from the global memory bank $\mathcal{M}$ using semantic similarity. Then we allocate the full-task memory to the orchestrator, and extract the subtask memories from the retrieved full-task memories and allocate the subtask memories to the corresponding task agents. 
As shown in \autoref{fig:framework-overview-sub}, for the vanilla \legomem{} variant, the orchestrator receives full-task memories that provide end-to-end workflows, while task agents are supplied with subtask memories that offer localized execution guidance. 
This design enables orchestrators to leverage prior trajectories for informed planning, agent capability grounding, and error recovery, while task agents improve their accuracy and efficiency in tool-use. 
As the task starts, the orchestrator receives the full memories and perform initial planning. Then, at each orchestration step, the orchestrator dynamically generates the next subtask using both the current state and retrieved full-task memories. 
The selected agent then executes the subtask with its allocated subtask memories, returning observations and summaries to update the orchestrator state. 
If progress stalls, the orchestrator can re-plan using memory as additional guidance. 
Through this loop, \legomem{} integrates past experiences to make more informed decision during planning and coordination, improving both reliability and efficiency of the multi-agent workflows.

\subsection{\legomem{} variants}
\label{sec:legomem-variants}

To explore the impact of subtask retrieval granularity in multi-agent systems, we compare three variants of \legomem{}: (vanilla) \legomem{}, \dynalegomem{}, and \queryrewritelegomem{}. 
These variants differ in how they store and retrieve subtask memories and allocate them to the task agents.

As discussed in \ref{sec:legomem-framework}, vanilla \legomem{} keeps a global procedural memory bank $\mathcal{M}$, and during inference, retrieves full-task memories using the task description and augment them to the orchestrator. Subtask memories are then extracted from these retrieved memories straightforwardly and are statically assigned to the relevant task agents. This approach is simple and efficient, and provides strong performance across teams. However, it may occasionally fail to surface relevant subtask memories for certain agents if the retrieved full-task memories differ in subtask structures from the current task. In such cases, even if the overall task appears similar, the subtask components may diverge. To address this, we implement two variants
%—\dynalegomem{} and \queryrewritelegomem{}—
that enable finer-grained subtask-level retrieval, improving task agent-level memory relevance:

\paragraph{\dynalegomem{}:}
As illustrated in \autoref{fig:dynamic-variant}, \dynalegomem{} performs subtask-level retrieval during execution. The orchestrator memory storage and retrieval remain the same as the vanilla version, while the system maintains per-agent subtask memory banks segmented from the global memory bank. When the orchestrator generates a subtask $s_t$ for an agent $A_t$, we compute its embedding $\phi(s_t)$ and query the agent’s memory bank $\mathcal{M}_{A_t}$ to retrieve only the most relevant past subtask traces. This just-in-time retrieval provides more precise execution guidance for task agents and reduces noise from irrelevant memories.

\paragraph{\queryrewritelegomem{}:}
While \dynalegomem{} performs just-in-time retrieval at each orchestration step, it incurs repeated subtask embedding and retrieval during execution. \queryrewritelegomem{} shifts this to the planning stage using query rewriting~\cite{ma2023query, lidmqr}. As shown in Figure~\ref{fig:query-rewrite-variant}, after retrieving full-task memories, a query rewriter LLM $\psi$ uses the memories to generate a draft plan for the new task $\pi'_{\text{draft}} = \{s'_1, s'_2, \dots, s'_n\}$ consisting of rewritten subtasks. Each $s'_j$ is then embedded via $\phi(s'_j)$ and used to retrieve relevant subtask memories from the corresponding agent’s memory bank $\mathcal{M}_{A_j}$ before task execution starts. This approach preserves the fine-grained retrieval benefits of \dynalegomem{} while avoiding repeated queries at runtime, enabling more efficient execution and smoother orchestration.

Interestingly, our experiments show that all three variants achieve similar overall performance in full memory settings, demonstrating the robustness across variants. Furthermore, our ablation study shows that \dynalegomem{} and \queryrewritelegomem{} outperform vanilla \legomem{} when only task agent-level memory is used and with small language model task agents. This indicates that fine-grained subtask retrieval may offer more relevant guidance to task agents and may be particularly beneficial in settings with weaker orchestrator support.

Together, the \legomem{} framework and its variants provide a general and modular approach to procedural memory for multi-agent LLM systems, enabling both orchestrators and task agents to learn from and reuse prior task executions. In the following section, we empirically evaluate these variants across different agent team configurations and memory settings.

%%%%%%%%%%%%%%%%%%%%%%%%%%%%% MAIN TABLE %%%%%%%%%%%%%%%%%%%%%%%%%%%%%
\begin{table*}[t]
\centering
\caption{Performance comparison across memory variants, task levels, and multi-agent teams. Results show mean success rates across different \legomem{} variants compared with baseline methods, each data-point is averaged over three random seeds.}
\label{tab:main_results}
\setlength{\tabcolsep}{3.5pt}
\begin{tabular}{lcccccccccccc}
\toprule
 & \multicolumn{4}{c}{\textbf{LLM team}} & \multicolumn{4}{c}{\textbf{Hybrid (LLM + SLM) team}} & \multicolumn{4}{c}{\textbf{SLM team}} \\
 & Level 1 & Level 2 & Level 3 & \textbf{Overall} & Level 1 & Level 2 & Level 3 & \textbf{Overall} & Level 1 & Level 2 & Level 3 & \textbf{Overall} \\
\midrule
\rowcolor{gray!24}
\textbf{Baseline methods} & & & & & & & & & & & & \\
No memory & 49.31 & 58.52 & 33.33 & 45.83 & 45.14 & 48.89  & 16.95 & 35.31 & 36.81 & 34.81 & 7.34 & 24.78 \\
Synapse & \textbf{59.72} & \textbf{75.56} & 43.50 & 58.11 & 46.53 & \textbf{68.15} & 29.94 & 46.49 & 36.81 & 42.22 & 20.90 & 32.24\\
AWM & 54.17 & 58.52 & 35.03 & 48.03 & 43.75 & 55.56 & 18.64 & 37.50 & 35.42 & 36.30 & 12.99 & 26.97 \\
\rowcolor{gray!24}
\textbf{Our methods} & & & & & & & & & & & &\\
\legomem{} & 57.99 & 73.33 & \textbf{47.46} & \textbf{58.44} & \textbf{49.31} & 62.22  & 36.16 & 48.03 & \textbf{38.89} & \textbf{54.07} & 25.42 & \textbf{38.16} \\
\dynalegomem{} & 56.25 & \textbf{75.56} & 43.79 & 57.12 & 44.44 & 65.93  & 36.16 & 47.59 & \textbf{38.89} & 50.37 & \textbf{27.12} & 37.72 \\
\queryrewritelegomem{} & 54.17 & 72.59 & 42.94 & 55.26 & 47.22 & 66.67 & \textbf{40.11} & \textbf{50.22} & 36.81 & 48.89 & 26.55 & 36.40\\
\bottomrule
\end{tabular}
\end{table*}

\begin{figure*}[t]
    \centering
    \includegraphics[width=\linewidth]{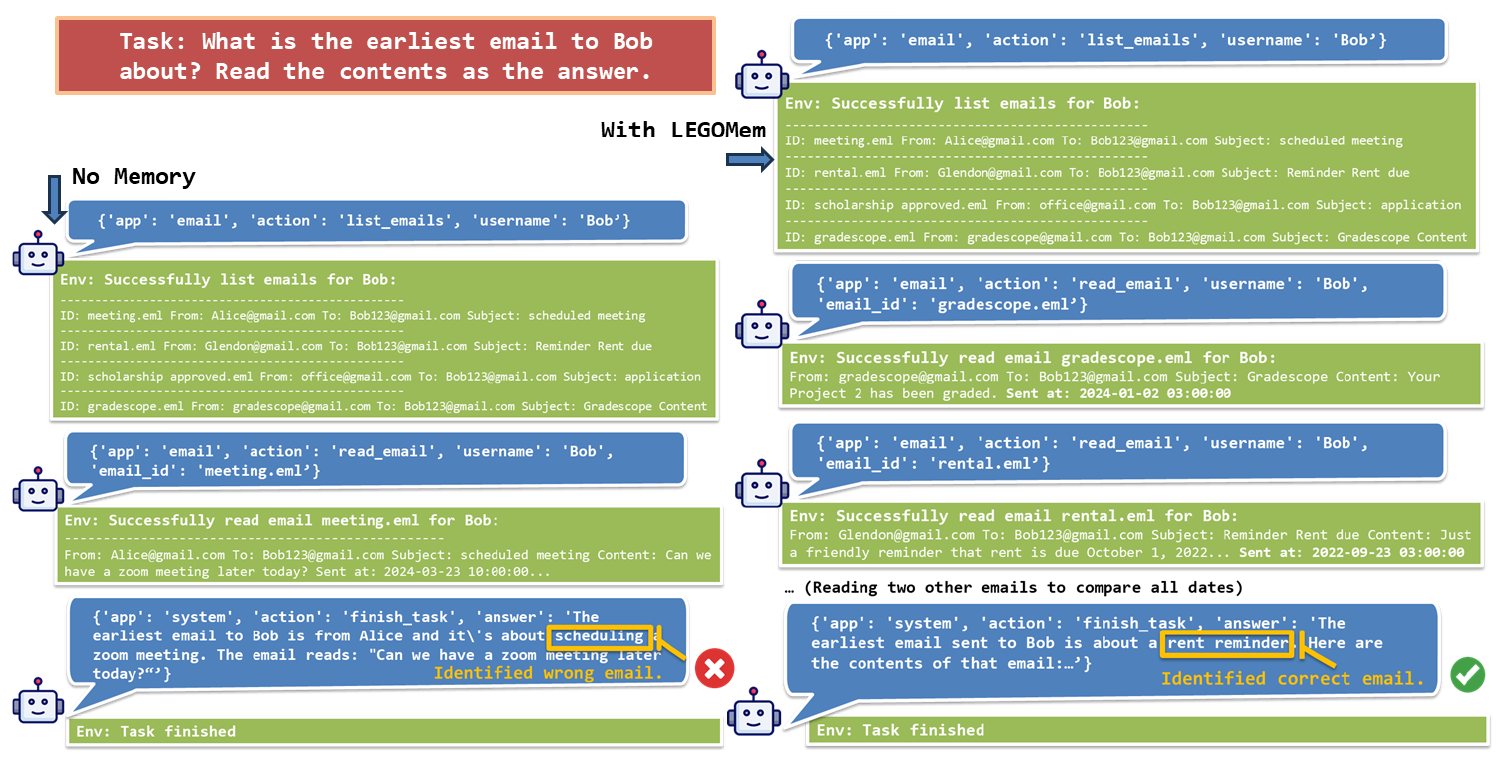}
    \caption{Qualitative example of agent execution with and without memory. The memory-less team fails to identify the earliest email due to incomplete planning, stopping after reading the first email, while the team with \legomem{} systematically reads to obtain and compare all email timestamps, producing the correct answer.}
    \label{fig:qualitative_example}
\end{figure*}

%%%% EXPERIMENT SECTION %%%%
\section{Experiments}
We evaluate \legomem{} on the OfficeBench benchmark, comparing its variants with strong baselines across LLM-only, hybrid, and SLM-only multi-agent teams. Beyond overall performance, we conduct ablations on memory placement, retrieval strategies, and representation formats to analyze the contributions of different design choices. Our results show that \legomem{} consistently improves task success rates across team configurations, and that memory design, particularly the placement of orchestrator memory, plays a central role in enabling effective multi-agent coordination.

\subsection{Experimental setup}
\subsubsection{Dataset and metrics}
We evaluate the agents on the OfficeBench, which consists of multi-step office automation tasks with varying levels of complexity. We split the 300 tasks into training (148 instances, for memory curation) and test (152 instances, for evaluation) sets. Tasks span three difficulty levels: Level 1 (single application), Level 2 (two-application), and Level 3 (multi-application workflows). 

The evaluation metric is the \textbf{success rate}, i.e. the percentage of tasks solved correctly. The success of a task is evaluated programmatically according to the final state of the environment, including exact match or fuzzy keyword match of the final outputs and expected outputs (e.g., correctly updated spreadsheet entries, calendar events, emails sent and received, and question answering).

\subsubsection{Implementation details}
We experiment with three team configurations with agents of different sizes and capabilities:
\begin{itemize}
    \item \textbf{LLM team}: for the full LLM team, we use GPT-4o~\cite{hurst2024gpt} for both the orchestrator and task agents
    \item \textbf{Hybrid (LLM + SLM) team}: GPT-4o for the orchestrator, and GPT-4o-mini for the task agents
    \item \textbf{SLM team}: GPT-4o-mini~\cite{hurst2024gpt} for all components
\end{itemize}
Additionally, for memory storage and retrieval, we use the OpenAI text-embedding-3-large model for embedding the task descriptions, and the FAISS library~\cite{douze2024faiss} for the vector database. For the OCR app, we use the Phi-3.5-mini model~\cite{abdin2024phi} as the vision language model for image parsing.

We compare the \legomem{} variants with three baselines: (i) \textbf{No memory}, and two state-of-the-art methods on procedural memory for workflow automation (ii) \textbf{Synapse}, which augments agents with semantically similar memories using raw action sequences and full trajectories, and (iii) \textbf{AWM}, which augments agents with summarized subtask memories extracted from full trajectories. 

\subsubsection{Memory curation and agent inference details}
Memory construction uses the 148 training tasks, where we first run the full LLM agent team without memory, and filter for successful trajectories and extracted 93 full task memories from the successful trajectories. For the \legomem{} variants, we further extracted 250 subtask memories for the task agents from the 93 full task memories. 
Both Synapse and AWM focus on single-agent systems; for a fair comparison we use the same 93 successful trajectories, and adapt both baselines to the multi-agent team, augmenting the memories to both orchestrators and task agents. For Synapse, we augment both orchestrators and task agents with the full trajectories. For AWM, we cluster the successful trajectories, to extract and consolidate subtask memories from each cluster, and during inference, we augment the task agents with their corresponding extracted subtask memories and augment the orchestrator with a list of extracted subtask memories. 
For all variants, we use 5 memories for orchestrator and 3 memories for each task agent from the successful trajectories.

%%%%%%%%%%%%%%%%%%%%%%%%%%%%% PLACEMENT TABLE %%%%%%%%%%%%%%%%%%%%%%%%%%%%%
\begin{table*}[t]
\centering
\caption{Comparing performance with various memory placement mechanism across \legomem{} variants.}
\label{tab:memory_placement}
\begin{tabular}{lcccccccc}
\toprule
 & \multicolumn{4}{c}{\textbf{LLM variants}} & \multicolumn{4}{c}{\textbf{Hybrid (LLM + SLM) variants}} \\
 & Level 1 & Level 2 & Level 3 & \textbf{Overall} & Level 1 & Level 2 & Level 3 & \textbf{Overall}\\
\midrule
\rowcolor{gray!20}
\textbf{Orchestrator + Agent memory} & & & & & & & & \\
\legomem{} & \textbf{57.99} & 73.33 & \textbf{47.46} & \textbf{58.44} & \textbf{49.31} & 62.22  & 36.16 & 48.03 \\
\dynalegomem{} & 56.25 & 75.56 & 43.79 & 57.12 & 44.44 & 65.93  & 36.16 & 47.59 \\
\queryrewritelegomem{} & 54.17 & 72.59 & 42.94 & 55.26 & 47.22 & 66.67 & \textbf{40.11} & \textbf{50.22}  \\
\rowcolor{gray!20}
\textbf{Orchestrator memory (planning) + Agent memory} & & & & & & & & \\
\legomem{} &  54.86 & \textbf{76.30}  & 35.03 & 53.51 & 45.14 & 63.70 & 30.51  & 44.96 \\
\dynalegomem{} &  54.86 & 73.33 & 41.81 & 55.26 & 46.53 & 64.44 & 32.77  & 46.49  \\
\queryrewritelegomem{} &  51.39 & 70.37  & 42.94 & 53.73 & 49.31 & 59.26 & 35.59  & 46.93 \\
\rowcolor{gray!20}
\textbf{Orchestrator memory} & & & & & & & & \\
\legomem{} &  51.39 & 74.07 & 38.98 & 53.29 & 45.83 & \textbf{68.89} & 32.77  & 47.59 \\
\rowcolor{gray!20}
\textbf{Task Agent memory} & & & & & & & & \\
\legomem{} & 50.00 & 63.70 & 38.98 & 49.78 & 44.44 & 46.67  & 19.21 & 35.31 \\
\dynalegomem{} &  49.31 & 62.96  & 38.98 & 49.34 & 47.22 & 55.56 & 23.16 & 40.35 \\
\queryrewritelegomem{} &  54.86 & 66.67  & 35.03 & 50.66 & 44.44 & 54.81 & 24.29  & 39.69 \\
\rowcolor{gray!20}
\textbf{No memory} & & & & & & & & \\
No memory & 49.31 & 58.52 & 33.33 & 45.83 & 45.14 & 48.89  & 16.95 & 35.31 \\
\bottomrule
\end{tabular}
\end{table*}

%%%%%%%%%%%%%%%%%%%%%%%%%%%%% REASONING TABLE %%%%%%%%%%%%%%%%%%%%%%%%%%%%%
\begin{table*}[t]
\centering
\caption{Comparing memory with and without reasoning across different \legomem{} variants.}
\label{tab:memory_representation}
\begin{tabular}{lcccccccc}
\toprule
 & \multicolumn{4}{c}{\textbf{LLM team}} & \multicolumn{4}{c}{\textbf{Hybrid (LLM + SLM) team}} \\
 & Level 1 & Level 2 & Level 3 & \textbf{Overall} & Level 1 & Level 2 & Level 3 & \textbf{Overall} \\
\midrule
\rowcolor{gray!24}
\textbf{No reasoning} & & & & & & & & \\
\legomem{} & 54.17 & \textbf{75.93} & 43.22 & 56.36 & 48.61 & \textbf{68.15}  & 36.72 & 49.78 \\
\dynalegomem{} & 56.94 & 73.33 & 44.63 & 57.02 & \textbf{50.00} & \textbf{68.15}  & 36.72 & \textbf{50.22} \\
\queryrewritelegomem{} & 48.61 & 69.63 & \textbf{48.02} & 54.61 & 37.50 & 65.19  & 36.16 & 45.18  \\
\rowcolor{gray!24}
\textbf{With reasoning} & & & & & & & & \\
\legomem{} & \textbf{57.99} & 73.33 & 47.46 & \textbf{58.44} & 49.31 & 62.22  & 36.16 & 48.03 \\
\dynalegomem{} & 56.25 & \textbf{75.56} & 43.79 & 57.12 & 44.44 & 65.93  & 36.16 & 47.59 \\
\queryrewritelegomem{} & 54.17 & 72.59 & 42.94 & 55.26 & 47.22 & 66.67 & \textbf{40.11} & \textbf{50.22} \\
\bottomrule
\end{tabular}
\end{table*}

\subsection{Main results}
\autoref{tab:main_results} presents the main experiment results, comparing the performance of \legomem{} with baseline methods across different task levels and agentic team configurations. 

Across all scenarios and agent team configurations, \legomem{} variants consistently outperform baseline methods in terms of overall success rate. All three \legomem{} variants show similar, consistent performance, with the vanilla \legomem{} variant being lightweight while achieving the best overall performance. The performance improvement shows the effectiveness of modularized memory representations and allocation for multi-agent systems.
Compared with memory-less teams, \legomem{} improves overall task success rate by $+12.61\%$, $+12.72\%$, and $+13.38\%$ absolute points on LLM, Hybrid and SLM teams, respectively.

Importantly, \legomem{} enables smaller models to close the gap with, and sometimes outperform, larger ones. For example, the Hybrid team with \queryrewritelegomem{} achieves $50.22\%$, surpassing the memory-less LLM team ($45.83\%$). Likewise, a full SLM team with vanilla \legomem{} ($38.16\%$) outperforms the Hybrid team without memory ($35.31\%$). While Synapse remains competitive in LLM teams, reflecting the ability of LLMs to interpret raw procedural traces, its effectiveness is less consistent for Hybrid and SLM teams. In contrast, \legomem{} maintains strong performance across all team settings, highlighting the importance of modularized procedural memory for enabling efficient, smaller-model teams.

To better illustrate the effect of memory on agent behavior, \autoref{fig:qualitative_example} presents a qualitative case study. Without memory, the agent fails to identify the earliest email due to incomplete planning, stopping after reading only the first entry. With \legomem{}, the agent systematically reads and compares all emails, correctly identifying the earliest one. This example highlights how \legomem{} improves reasoning consistency and task completeness beyond what is reflected in aggregate success rates.

\begin{figure*}[t]
    \centering
    \begin{subfigure}{0.48\textwidth}
        \centering
        \includegraphics[width=\linewidth]{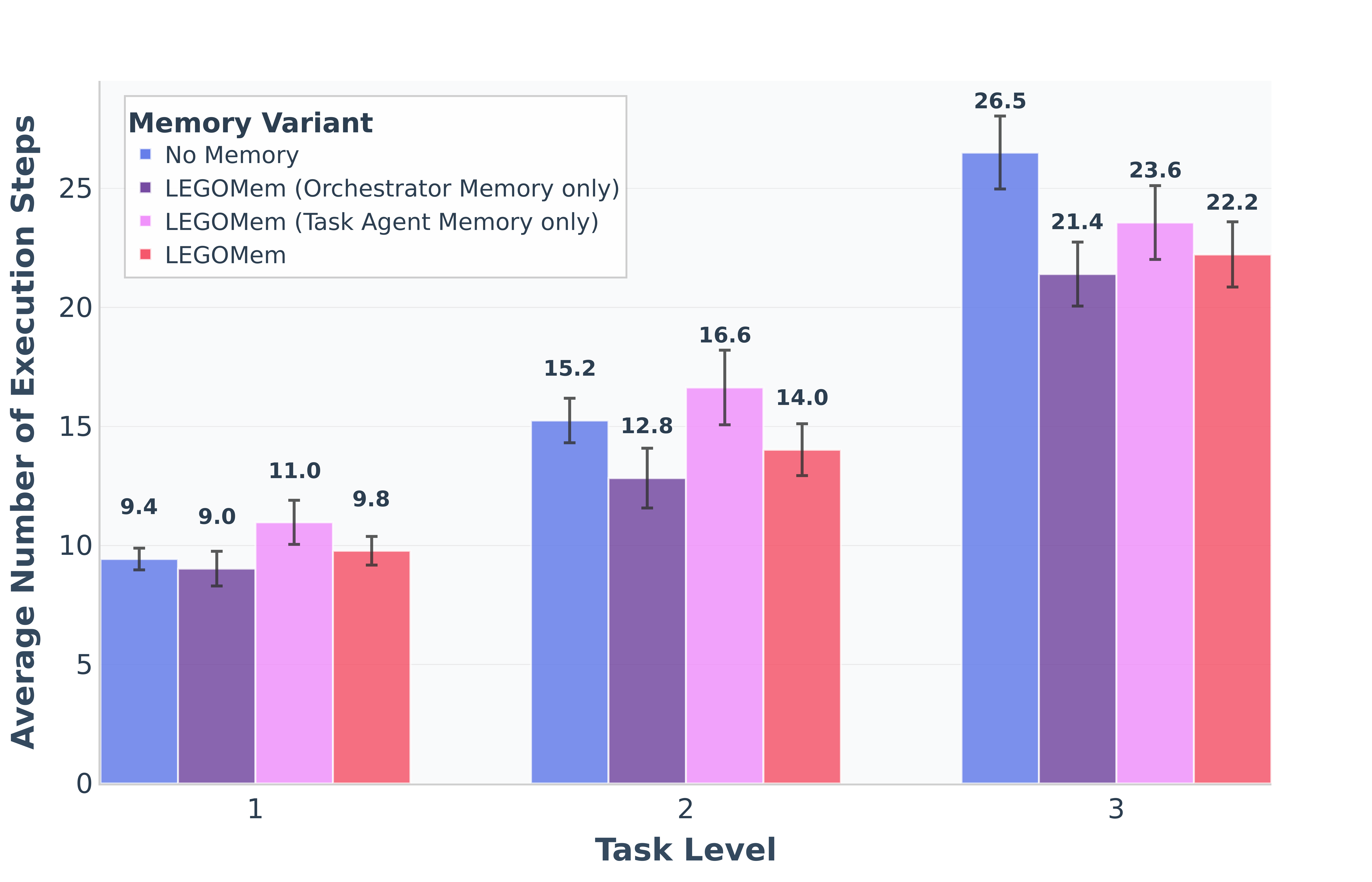}
        \caption{Average execution steps across task levels}
        \label{fig:avg-execution-steps}
    \end{subfigure}
    \hfill
    \begin{subfigure}{0.48\textwidth}
        \centering
        \includegraphics[width=\linewidth]{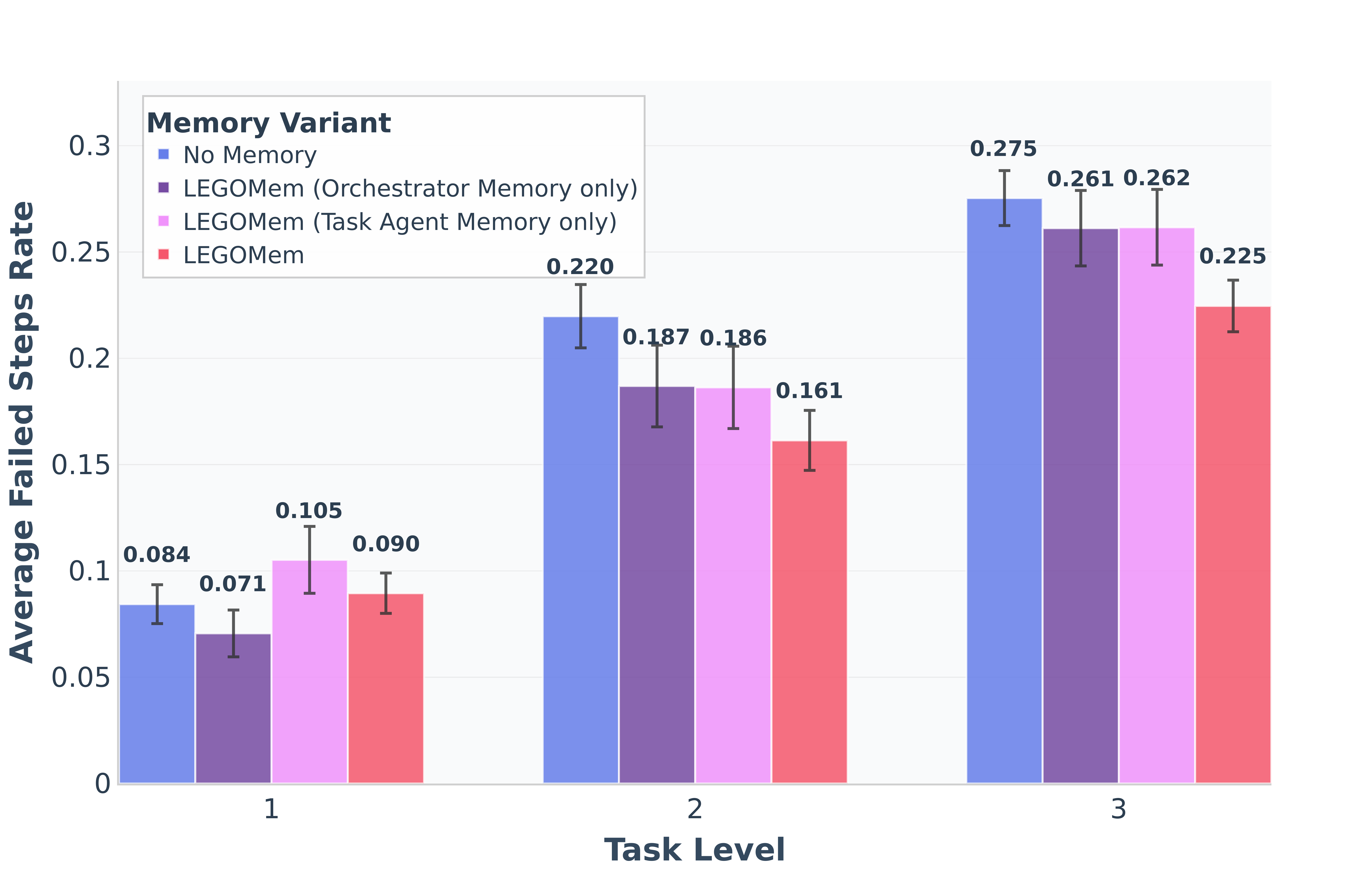}
        \caption{Average failed steps rate across task levels}
        \label{fig:avg-failed-steps}
    \end{subfigure}
    \caption{Ablations study: execution steps comparison for different \legomem{} memory placement for LLM teams. (a) shows that \legomem{} variants reduce the number of execution steps required, with up to 16.2\% reduction for Level 3 tasks. (b) shows lower failure rates of steps, indicating more reliable task execution with procedural memory.}
    \label{fig:execution-efficiency}
\end{figure*}

\subsection{Ablations experiments}
This section investigates how different memory retrieval, allocation, and placement strategies affect the performance of \legomem{}.

\subsubsection{Memory retrieval, allocation, and placement}
\autoref{tab:memory_placement} summarizes our ablation results across different memory retrieval variants, memory allocation strategies, and memory placement settings.

\paragraph{Memory retrieval} The three subtask memory retrieval strategies—vanilla \legomem{}, \dynalegomem{}, and \queryrewritelegomem{}—all perform robustly and achieve similar overall success rates. While dynamic retrieval enables more targeted allocation and query rewriting improves robustness to subtask phrasing variations, these differences are modest compared to the impact of memory placement and allocation strategy.

In the task-agent-only memory setting, both \dynalegomem{} and \queryrewritelegomem{} outperform vanilla \legomem{} by 4–5\% on average in the Hybrid team where task agents are smaller models and agent-level memory plays a more critical role. These results highlight the advantage of fine-grained subtask retrieval in providing more relevant and contextual guidance to task agents, especially when global planning signals are weaker. We hypothesize that the similar overall performance of all three variants in full-memory settings may be due to the strength of the orchestrator memory, where the orchestrator receives the complete trajectory of prior solutions, compensating for weaker task agent execution by enabling better task decomposition and delegation. 

Overall, these findings demonstrate the flexibility of the \legomem{} framework: even the lightweight vanilla variant performs competitively, while more advanced variants offer additional benefits in settings that demand finer-grained memory retrieval.

\paragraph{Memory allocation} Regarding memory allocation, we find that joint allocation of orchestrator and task agent memory (Orchestrator + Agent memory variant) yields the strongest overall results, with orchestrator memory supporting effective planning, task decomposition and subtask orchestration, and task agent memory enabling execution-level precision. Orchestrator memory emerges as essential: when memory is removed from the orchestrator and provided only to task agents (Task Agent memory variant), performance drops noticeably.

\paragraph{Memory placement} Looking at memory placement, even when restricted to the planning and replanning stages, orchestrator memory still improves over task-agent-only variants, confirming its central role in guiding high-level planning and task decomposition. Finally, Task-Agent-only memory while facilitating more accurate tool use and outperforming the no-memory baseline, remains less effective than orchestrator-level memory -- indicating that local memory without global coordination is insufficient.

\subsubsection{Effectiveness of adding reasoning in memory}
We also examine whether augmenting procedural memories with lightweight reasoning improves performance. As shown in \autoref{tab:memory_representation}, the differences are minor: overall scores change by less than two points across variants and team types. For example, vanilla \legomem{} improves slightly on LLM teams ($56.36\%\rightarrow58.44\%$) but decreases on Hybrid teams ($49.78\%\rightarrow48.03\%$). These results suggest that \legomem{} is robust, with its modularized structure already providing sufficient procedural guidance without additional reasoning steps.

\subsubsection{Effectiveness of memory on execution steps and failure rates}
As an additional ablations study, \autoref{fig:execution-efficiency} compares the average number of execution steps taken by the agent with different memory placement variants and the step failure rate (due to wrong tool-use actions issued) per task for the LLM team. As shown in \autoref{fig:avg-execution-steps} Compared to the no memory variant, the agents equipped with \legomem{} can reduce the number of execution steps required to complete the tasks, for example, a -16.2$\%$ drop from an average of 26.5 to 22.2 steps for Level~3 tasks. The task memory only variant where we remove the orchestrator memory required more steps to complete a task compared with the variant with orchestrator memory, due to the effectiveness of the orchestrator memory for improved planning. 

Similarly, \autoref{fig:avg-failed-steps} shows that \legomem{} reduces the average failure rate of agent steps. At Level~3, the failure rate decreases from 0.275 in the no-memory setting to 0.225 with \legomem{}. These results indicate that \legomem{} not only improves task success rates but also enables more efficient and reliable task execution.

In summary, our experiments show that \legomem{} consistently outperforms baselines methods, improving task success by over 12 absolute percentage points compared with memory-less teams. \legomem{} can enable smaller and hybrid teams to match or even surpass LLM-only teams, highlighting its value for efficient multi-agent configurations. Ablations reveal that, as one may expect, the memory placement strategy is critical: orchestrator memory is essential for effective planning, while subtask memory complements execution. Additional analysis also show reductions in execution steps required and per-step failure rates with \legomem{}. 

\section{Conclusion}

We introduced \legomem{}, a modular procedural memory framework for multi-agent systems that enables orchestrators and task agents to learn from prior task executions. By representing workflows as reusable memory units—split into full-task and subtask components—\legomem{} supports efficient task planning and execution through memory retrieval and allocation.
We implemented and evaluated three \legomem{} variants to explore the design space of memory retrieval and placement strategies. Across extensive experiments on workflow automation tasks, we show that \legomem{} significantly improves task success rates over memory-less and baseline methods, with orchestrator memory playing a critical role in planning and coordination, and memory can also benefit smaller agents, highlighting the flexibility and effectiveness of the framework.
Our work shows that integrating procedural memory into multi-agent systems enables more reliable and reusable solutions. Future work may explore continual learning also from failed past trajectories, and scaling \legomem{} to open-ended environments and tool ecosystems.

%%%%%%%%%%%%%%%%%%%%%%%%%%%%%%%%%%%%%%%%%%%%%%%%%%%%%%%%%%%%%%%%%%%%%%%%

%%% The acknowledgments section is defined using the "acks" environment
%%% (rather than an unnumbered section). The use of this environment 
%%% ensures the proper identification of the section in the article 
%%% metadata as well as the consistent spelling of the heading.

% \begin{acks}
% If you wish to include any acknowledgments in your paper (e.g., to 
% people or funding agencies), please do so using the `\texttt{acks}' 
% environment. Note that the text of your acknowledgments will be omitted
% if you compile your document with the `\texttt{anonymous}' option.
% \end{acks}

%%%%%%%%%%%%%%%%%%%%%%%%%%%%%%%%%%%%%%%%%%%%%%%%%%%%%%%%%%%%%%%%%%%%%%%%

%%% The next two lines define, first, the bibliography style to be 
%%% applied, and, second, the bibliography file to be used.

\bibliographystyle{ACM-Reference-Format} 
\bibliography{references}

%%%%%%%%%%%%%%%%%%%%%%%%%%%%%%%%%%%%%%%%%%%%%%%%%%%%%%%%%%%%%%%%%%%%%%%%

\newpage
\onecolumn
\appendix
\input{appendix}

%%%%%%%%%%%%%%%%%%%%%%%%%%%%%%%%%%%%%%%%%%%%%%%%%%%%%%%%%%%%%%%%%%%%%%%%

\end{document}

%% file: appendix.tex
\section{Prompts for Memory Curation}
In this section, we provide the detailed prompts for memory curation and the prompt for the query rewriting LLM.
\begin{Prompt}{\textbf{Memory Curation Prompt}}{}
From the following agent trajectory, generate memory that can be useful for future LLM agents' reference.

\# \textbf{Trajectory:}
$\{$\textbf{full\_trajectory}$\}$

\# \textbf{Example:}\\
$\{$start\_tag$\}$\\
$\{\{$
    "high\_level\_plan": "1. Check Bob's calendar availability for the specified time slot. 2. Add the meeting to Bob's calendar for 5\/17\/2024 from 10:30 a.m. to 11:00 a.m.",
    "subtasks": [
        $\{\{$
            "agent": "calendar\_agent",
            "description": "Check Bob's schedule on 5/17/2024 from 10:30 a.m. to 11:00 a.m to ensure there are no conflicts",
            "steps": "<think>I need to check Bob's existing calendar events to ensure no scheduling conflicts</think><action>$\{\{"app": "calendar", "action": "list\_events", "username": "Bob"\}$</action>",
            "observations": "No events found for Bob - calendar is available for the requested time slot"
        \}\},
        \{\{
            "agent": "calendar\_agent",
            "description": "Add a meeting to Bob's calendar on 5/17/2024 from 10:30 a.m. to 11:00 a.m",
            "steps": "<think>Since no conflicts were found, I can now create the new calendar event for Bob</think><action>$\{\{"app": "calendar", "action": "create\_event", "user": "Bob", "summary": "Meeting", "time\_start": "2024-05-17 10:30:00", "time\_end": "2024-05-17 11:00:00"\}\}$</action>",
            "observations": "Successfully created a new event in Bob's calendar for the specified date and time"
        $\}\}$
    ],
    "final\_answer": "The meeting has been successfully added to Bob's calendar on 5\/17\/2024 from 10:30 a.m. to 11:00 a.m.",
    "reflections": "Task completed successfully without any conflicts or errors. The calendar check confirmed availability, and the meeting was created with proper date/time formatting."
$\}\}$
$\{$end\_tag$\}$

\# \textbf{Instructions:}\\
Please analyze the trajectory and extract structured memory with clear thinking and well-formed actions. Use the following format for each subtask step:
<think>reasoning about what needs to be done and why this action is appropriate</think>
<action>{{precise tool call command in structured format}}</action>

The memory object should be formatted as follows:
\{\{
    "high\_level\_plan": "<a string that lists the high-level steps taken and which agent performs each subtask>",
    "subtasks": [
        \{\{
            "agent": "<copy the exact name of agent that performed the subtask>",
            "description": "<description of the subtask given by the orchestrator>",
            "steps": "<Copy the precise actions taken with think-action structure: <think>reasoning</think><action>${{tool\_call}}$</action>, repeat for each action. Omit some actions if there are too many similar commands (>10). Remove actions that yielded errors or were malformed.>",
            "observations": "<a very brief summary of the key observations from the function execution results>",
        \}\},
        ...
    ],
    "final\_answer": "<The final answer given by the orchestrator or answer agent>",
    "reflections": "<a concise summary that lists what was successful, what were specific failures, root cause of which action and how to avoid, if any>",
\}\}

\# \textbf{Rules to follow:}\\
1. Group together actions into subtasks if they are related and can be done together.\\
2. For each action in the steps field, use the think-action format with clear reasoning followed by structured tool calls.\\
3. When copying actions, remove function call IDs but keep the essential tool call structure.\\
4. Only include successful actions; omit actions that resulted in errors. If there are too many repeated similar actions, truncate and omit some, and if the action parameters (such as contents to write to a word document) are too long, you can summarize it.\\
5. Keep observations very concise but informative.\\
6. Do not include orchestrator coordination steps in the subtasks.\\
7. For the subtask steps field, use a string format with think-action pairs, not a list.\\

Follow the JSON format exactly to ensure it can be parsed automatically, and put the json object between the tags $\{$start\_tag$\}$
\# your json here
$\{$end\_tag$\}$ and do not use markdown.

\end{Prompt}

\begin{Prompt}{\textbf{Query Rewriting Prompt}}{}
Based on the following similar task examples, break down the new task into a step-by-step plan.

\#\# \textbf{Similar Task Examples:}
$\{$memory\_context$\}$

\#\# \textbf{New Task: }
$\{$task\_description$\}$

Please provide a numbered list of 3-5 high-level steps that would be needed to complete this task.
Focus on the main phases/subtasks, not detailed actions.

Format your response as a simple numbered list enclosed within <start> and <end> tags:

<start>
1. [First step]
2. [Second step]
3. [Third step]
...
<end>

Steps:
\end{Prompt}